\documentclass[journal]{IEEEtran}
\usepackage{color}
\usepackage{algorithm2e}
\usepackage{algorithmic}
\usepackage{pgfplots}
\usepackage{amsfonts}
\usepackage{amssymb}
\usepackage{mathtools}
\usepackage{bbm}
\usepackage{dsfont}
\usepackage{float,comment}

\ifCLASSINFOpdf
\else
\fi
\begin{document}
%
\title{SACT: Self-Aware Multi-Space Feature Composition Transformer for Multinomial Attention for Video Captioning}
%
%
%

\author{Chiranjib~Sur \\ 
		Computer \& Information Science \& Engineering Department, University of Florida.\\
		Email: chiranjibsur@gmail.com
}

%
%

\markboth{Journal of XXXX,~Vol.~XX, No.~X, AXX~20XX}%
{Shell \MakeLowercase{\textit{et al.}}: Bare Demo of IEEEtran.cls for IEEE Journals}
%

\maketitle

\begin{abstract}
Video captioning works on the two fundamental concepts, feature detection and feature composition. While modern day transformers are beneficial in composing features, they lack the fundamental problems of selecting and understanding of the contents. As the feature length increases, it becomes increasingly important to include provisions for improved capturing of the pertinent contents. In this work, we have introduced a new concept of Self-Aware Composition Transformer (SACT) that is capable of generating Multinomial Attention (MultAtt) which is a way of generating distributions of various combinations of frames. Also, multi-head attention transformer works on the principle of combining all possible contents for attention, which is good for natural language classification, but has limitations for video captioning. Video contents have repetitions and require parsing of important contents for better content composition. In this work, we have introduced SACT for more selective attention and combined them for different attention heads for better capturing of the usable contents for any applications. 
To address the problem of diversification and encourage selective utilization, we propose the Self-Aware Composition Transformer model for dense video captioning and apply the technique on two benchmark datasets like ActivityNet and YouCookII. 
\end{abstract}

\begin{IEEEkeywords}
classification, clustering, .
\end{IEEEkeywords}

%
\IEEEpeerreviewmaketitle

\section{Introduction} \label{section:introduction}

\begin{figure}[ht]
\vskip 0.2in
\begin{center}
\centerline{\includegraphics[width=\columnwidth]{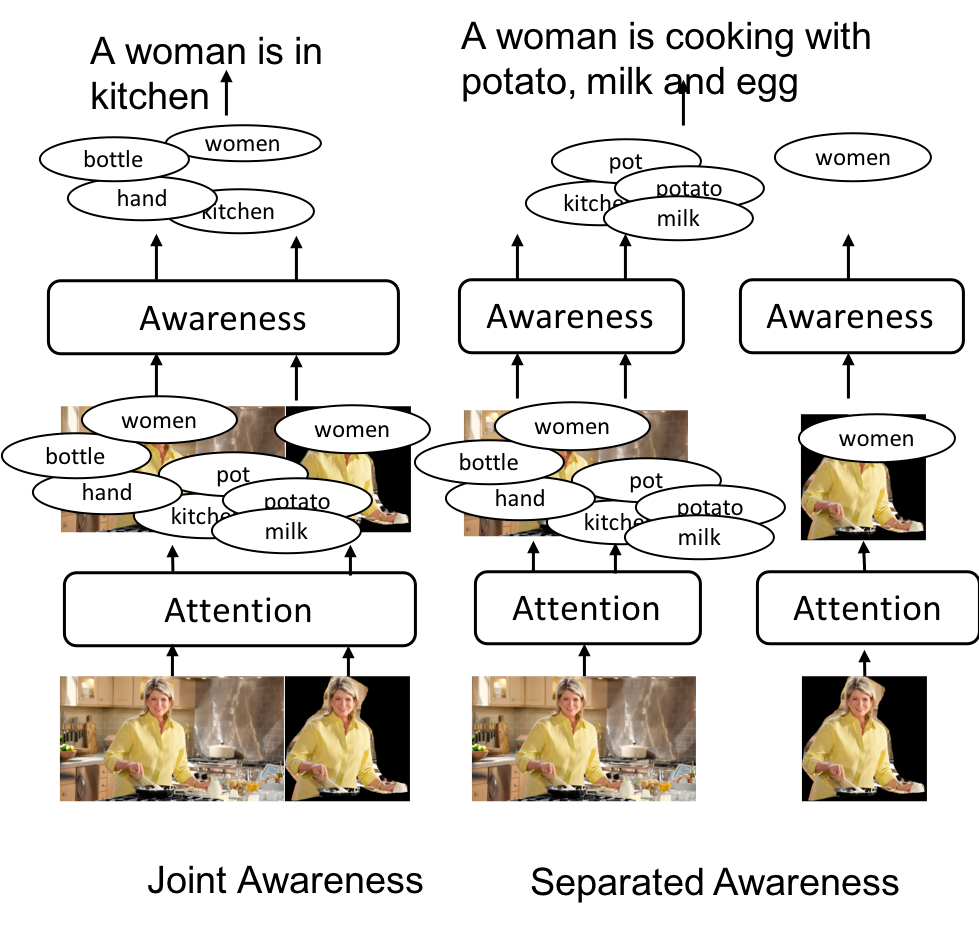}}
\caption{Defining Awareness Strategies. Joint Awareness has Joint Influence. While Separated Awareness have Exploration.}
\label{fig:aware}
\end{center}
\vskip -0.2in
\end{figure}

Video Captioning has gained a significant amount of attention as massive amounts of media content are created each day. Media help humans with instructions like video lectures, and also in day-to-day activities like preparing food and repairing. It is very important to develop the capability to comment and understand the different aspects of the video contents and integrate the understandings with the machines. 
Captioning is one way for humans to understand the content without watching it thoroughly. Captioning is a way of compressing media content, yet retaining the semantics. 
Recently, researchers concentrated on dense video captioning \cite{krishna2017dense}, which is a way of detecting and narrating events in the video with descriptive natural language compared to unstructured video summarization \cite{zhang2016video}. 
Recently, researchers worked on different captioning models through deep neural network-based methods \cite{vinyals2015show, vinyals2016show} and Convolutional Neural Network (CNNs) features \cite{simonyan2014very} for perfect descriptions of different events. Several methods are tried including Markov models and ontologies \cite{yu2013grounded, das2013thousand}, mean-pooling \cite{venugopalan2014translating, gan2017semantic}, recurrent nets \cite{venugopalan2015sequence}, and attention mechanisms \cite{yao2015describing, pan2017video}. 

Other works in language generation includes attention mechanism for machine translation \cite{bahdanau2014neural}, through cross-module attentions like temporal attention \cite{yao2015describing}, semantic attention \cite{gan2017semantic} or both \cite{pan2017video}, self-attention \cite{vaswani2017attention} .  Attention mechanism within a module through self-attention \cite{lin2017structured, paulus2017deep, vaswani2017attention} has improved performance considerably. 
\cite{shou2016temporal} introduced proposal candidates over video frames in a sliding window fashion.  Recent work in event detection include anchoring mechanism from object detection like  explicit anchoring \cite{zhou2018towards} and implicit anchoring \cite{buch2017sst}. 
The explicit anchoring methods accompanied with location regression yield better performance \cite{gao2017turn}.  \cite{zhou2018towards} detected long complicated events rather than actions. Some of the benchmark works included LSTM-YT \cite{venugopalan2014translating}, S2YT \cite{venugopalan2015sequence}, TempoAttn \cite{yao2015describing}, H-RNN \cite{yu2016video} and DEM \cite{krishna2017dense}. ProcNets \cite{zhou2018towards} worked on long dense event proposals with 1-D convolution. 
The video paragraph captioning \cite{yu2016video} generated sentence descriptions for temporally localized video events. 
\cite{das2013thousand} produced dense captions relying on a top-down ontology for the actual description.
Several other approaches like \cite{khan2011human, hanckmann2012automated, barbu2012video} introduced rule-based systems for objects and events in different tasks and scenarios, including complex rule-based system \cite{barbu2012video}. 
Speech has different mathematical representations and used in  \cite{guadarrama2013youtube2text, krishnamoorthy2013generating}.
Rule engineering with statistical models for fully \cite{guadarrama2013youtube2text, sun2014semantic} or weakly \cite{rohrbach2014coherent} supervised fashion were also introduced.
Larger datasets like YouTubeClips \cite{chen2011collecting} and TACoS-MultiLevel \cite{rohrbach2014coherent} are quite popular.

\begin{figure*}[ht]
\vskip 0.2in
\begin{center}
\centerline{\includegraphics[width=2\columnwidth]{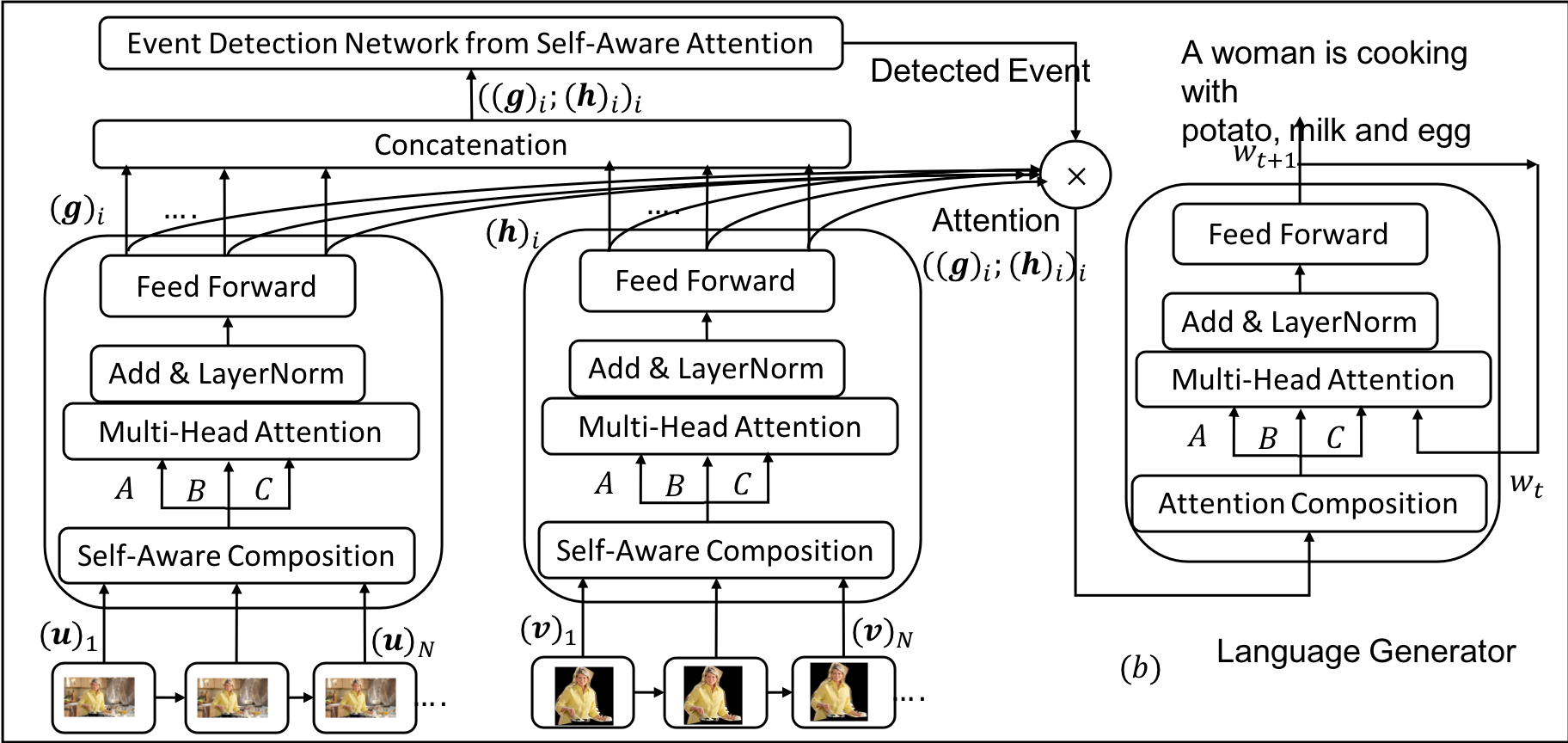}}
\caption{Overall Architecture of the Caption Generated Model with Self-Aware (Separated) Composition Transformer (SA$_s$CT). }
\label{fig:sact2}
\end{center}
\vskip -0.2in
\end{figure*}

Normally the event detector and the caption generator are trained alternatively \cite{ghanem2017activitynet} for fine-tuning the two models or independently or as an end-to-end \cite{krishna2017dense}. It is important for the semantic attention, generated in the transformer, to have direct influence on the event proposal. This way, the architecture will be more aware of the representation tensors. Adopting the event proposal decoder framework of \cite{zhou2018towards} and the end-to-end dense caption flow of \cite{zhou2018end}, we introduced a self-aware novel architecture framework that can create better performance for caption generation for videos. While traditional attention is dependent on weighted feature extraction, we involve a learning based filtering process that can decode the multi-frame content and compose better attention. 

Previous works, mainly, focused on the principle that the neural weights will detect the events and then capture different aspects of the video contents through encoding the whole sequence. But in reality, the weights are not sensitive to contents, instead they approximate. To make the model learn more about the event and better representation, we have proposed a novel architecture called Self-Aware Composition Transformer (SACT). It can counter the limitation of dealing with multiple frames and establish the perfect content for attention generation. Awareness is identification of important attentions. 
Most of the previous works concentrated on Multiple-Attentions (MA). MA adds the consistency of the feature (adjacency) relationships. That means if we have stream of features $ \{ \textbf{I}_1, \ldots,  \textbf{I}_n \}$ for $ n $ number of video frames, then MA can be denoted as $ f_{_{MA}} (\textbf{I}_1, \ldots, \textbf{I}_n)  = W_{MA} \phi (\textbf{I}_1, \ldots, \textbf{I}_n) $. But, we introduced SACT and Multinomial Attention (MultAtt) to mitigate the deficiencies present in MA. 
The main advantage of SACT architecture is that it has Multinomial Attention (MultAtt) instead of Multiple Attention (MA). MultAtt considers the distribution of the various combinations of frames, considering the success of the frames to be more relevant and important. MultAtt can be denoted as the following equation, 
$ f_{_{M_uA}} (\textbf{I}_1, \ldots, \textbf{I}_n)  = W_{M_uA} \sum\limits_{i=1}^{n} \beta_i \textbf{I}_i = W_{M_uA} \sum\limits_{i=1}^{m} \beta_i \textbf{I}_i  $ and $m << n$ so we have lot of $\beta_i = 0$ for different $i$. $\beta_i$ is the estimated variable depending on the utility of the features $i$. The estimation of $\beta_i$ for all $i \in \{1,\ldots,n\}$ happens for all the heads in the multiple-head attention transformer framework, creating ample opportunities for different feature combinations for the application.    
In video datasets, the feature relationships between different frames are related with common features, while feature variations are important for understanding the feature space. Both of these (variation and  common) are important. Subsequently, we have utilized both aspects of the multiple frame video for feature attention. We define the notion of MultAtt as a mode to capture the multiple frames based on topologically related sequence data. 

\begin{figure*}[ht]
\vskip 0.2in
\begin{center}
\centerline{\includegraphics[width=2\columnwidth]{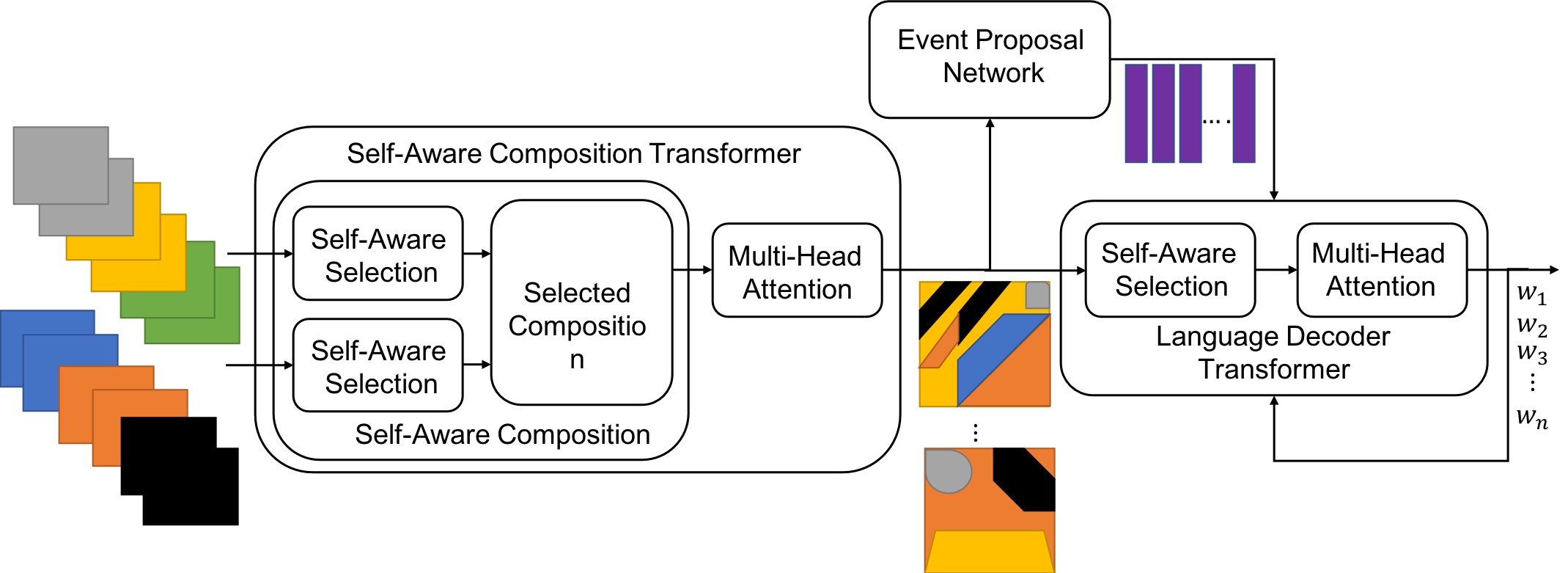}}
\caption{Architectural Details of Self-Aware Composition Transformer (SACT) Model for Video Captioning Considering Multiple Source of Features Like Lower Level Convolution Layers and Optical Flow.}
\label{fig:an1}
\end{center}
\vskip -0.2in
\end{figure*}

Video captioning aims to generate natural language based descriptions for all possible events that can be detected in an untrimmed video. Dense video captioning aims at detecting events and narrating them. However, event detection in a video and narrating the incidents were defined as separate problems. \cite{zhou2018end} defined a masked transformer approach where they provided the scope of multi-headed attention for video captioning application. It works on the principle of attention combination with all possible presence. However, we propose a novel SACT framework that restricts features and generates much better attention and is evident from the experiments.  
We adopted the proposal module of \cite{zhou2018towards, zhou2018end} for detecting the events in the video. Also, it provides a baseline for comparison and helps in understanding the impact of self-awareness on generating captions. Self-awareness of the image features will also create better representation for the attentions with limited and important contents. This will directly impact the generated captions. Due to the high dimension of the temporal contents with redundancy, self-awareness will help generate a better sub-space for representation. This will generate better attention for the framework. 

The rest of the document is arranged with 
architectural intricacies and details of implementation in Section \ref{section:architecture}, analysis of the experiments in Section \ref{section:results} and concluding remarks in Section \ref{section:conclusion}. 

The main contributions of this work. 1)  a novel transformer framework called Self-Aware Composition Transformer (SACT) with feature understanding at the frame level for better video captioning application 2) the notion of Multinomial Attention (MultAtt) for better attention, in comparison to the previous approaches  3) our approach is a much more realistic way of solving the multiple frame based problems, prevents contamination with excessive frames and less burden on the model to capture the content  4) outperformed some of the previous works in video captioning applications.


\section{Architecture} \label{section:architecture}
Self-Aware Composition Transformer (SACT) works on the principle of identification of the different contents of the feature that are useful, with possible attention created from the useful ones. This will create a difference between a traditional transformer and SA$_*$CT architecture with $* = s,j$. 
Multiple attention is problematic for models, hence, the model must be able to identify the usefulness of the features, that are more important than the other. Awareness is identification of important attentions.
It helps in fine-grained visual content processing through multiple layers of reason-processing to generate high quality captions. 
It can help in creating Multinomial Attentions that are defined by the various combinations of frames. The Self-Awareness is learnt considering the success of the frames that is considered relevant and important. We have introduced a novel transformer framework that can identify the important frame(s) in a sequence and can parse them as it is for better propagation of information. While we can create a framework that can parse the relevant important information, it is also important to deal with the situations that occupy different feature spaces. Different feature spaces provide different stories of the same sequence. We have provided a technique that can identify different important features and provides the required composition for attention.  
The details of the SACT architecture can be sub-divided into different tasks and these are performed for better understanding of the underlying object feature space and their compositional characteristics. 

\subsection{Detailed Transformer Characteristics}
Transformer generates multi-headed attentions through scaled dot-product of all possible combinations of the features \cite{vaswani2017attention, lee2018set, devlin2018bert, sur2020rbn}. A few variants are like set transformer \cite{lee2018set}, evolved transformer \cite{so2019evolved}. However, the attention generation is influenced by the closeness of the features in the multiple-frame sequence. This is a serious drawback. We solved this problem through our proposed network and creating awareness in the network. It is an inevitable procedure for machines to learn about the importance of the sequences. Mathematically, the transformer can be denoted as the following equation,
\begin{equation}
\begin{split}
 f & = \psi(\textbf{W}_p(\textbf{u};\textbf{v}), \textbf{W}_q(\textbf{u};\textbf{v}), \textbf{W}_r(\textbf{u};\textbf{v})) \\
   & = \text{softmax} \left( \frac{\textbf{W}_p(\textbf{u};\textbf{v}) * \textbf{W}_q(\textbf{u};\textbf{v})}{\sqrt{d}} \right) \textbf{W}_r(\textbf{u};\textbf{v})
 \end{split}
\end{equation}
With multiple-head attention, the transformer framework takes the following form, 
\begin{equation}
\begin{split}
 f & = \psi_N(\textbf{W}_p(\textbf{u};\textbf{v}), \textbf{W}_q(\textbf{u};\textbf{v}), \textbf{W}_r(\textbf{u};\textbf{v})) \\
   & = [\text{head}_1, \text{head}_2, \ldots, \text{head}_h] \textbf{W}
 \end{split}
\end{equation}
where we define $\text{head}_h$ as,
\begin{equation}
 \text{head}_h = \psi(\textbf{W}_{p_h}(\textbf{u};\textbf{v}), \textbf{W}_{q_h}(\textbf{u};\textbf{v}), \textbf{W}_{r_h}(\textbf{u};\textbf{v}))
\end{equation}
We are more acquainted to see this framework with $q$, $k$ and $v$ as query, key and value like transformer framework \cite{vaswani2017attention}. But, for self-aware composition transformer scenario, we can regard $(\textbf{u};\textbf{v})$ as composer for $q$, selector for $k$ and amplifier $v$. The $\textbf{W}_p$ has the  composition property, $\textbf{W}_q$ introduces the scaled composition based on what the network thinks is the most useful frames, $\textbf{W}_r$ helps in amplifying the contents. $\textbf{W}_p$ and $\textbf{W}_q$ generates the right set of combination attention. Multiple heads make sure that the approximation of the weighted transformation is compensated with multiple representations of the same semantic concept. The last $\textbf{W}_r$ amplifies the contents. This is evident from usage of softmax layer which provides a non-linear transformation to the content. For improved influence of the softmax layer as representation,  $\textbf{W}_r$ helps provide the necessary amplification of the semantic composition.

\subsection{Self-Aware Composition Attention} %
Self-Aware Composition Attention (SACA) is the added functionality that is provided in the transformer network for capturing multinomial attention. Transformer attention is influenced with adjacency of tensors. To avoid such scenarios, we have influenced the attention with selection based on awareness. Awareness is the capability to identify the important feature subset. 
Though multi-head attention helps in capturing different aspects, they lack coherence and uncontrolled contamination of irrelevant features. The reason is that there are not enough procedures that can contribute to the contents of the frames. Self-Aware Composition Attention provides a solution for that problem. 
For multiple data frame based problems, it is not only important to identify the regions of interest, but also create scope for awareness of the frames that are useful. This will reduce the contamination and can potentially help in enhanced composition of the representations. While self-attention uphold different overview of the whole content, we proposed a model that can be sensitive to the content and transfer to completely new framework and different narratives, which the model can identify for captions generation. The SACA function $ \Phi(.) $ can be denoted as the following,
\begin{equation} \label{eq:SACA1}
  \Phi(\textbf{u},\textbf{v}) = ( \omega_u * \textbf{u}  \text{ } ; \text{ }  \omega_v * \textbf{v} )
\end{equation}
where we define the awareness selector  $\omega_u, \text{ } \omega_v,\text{ }  \omega_{(u,v)} \in \mathbb{R}^{d}$ for $d$ features as, 
\begin{equation} \label{eq:SACA2}
 \omega_u = \sigma ( \textbf{Z}_{_{u_1}} \rho( \textbf{Z}_{_{u_2}} \textbf{u} + \textbf{Z}_{_{u_3}} \overline{\textbf{u}} ) )
\end{equation} 
\begin{equation} \label{eq:SACA3}
 \omega_v = \sigma ( \textbf{Z}_{_{v_1}} \rho( \textbf{Z}_{_{v_2}} \textbf{v} + \textbf{Z}_{_{v_3}}\overline{\textbf{v}}) )
\end{equation} 
\begin{equation} \label{eq:SACA4}
 \omega_{(u,v)} = \sigma ( \textbf{Z}_{_{v_1}} \rho( \textbf{Z}_{_{v_2}} \textbf{(u,v)} + \textbf{Z}_{_{v_3}}\overline{\textbf{(u,v)}}) )
\end{equation}
where we have $\sigma (.)$ and $\rho (.)$ as two functional approximation. $ \omega $ is a sparse matrix with higher weight-age on the useful feature spaces. $ \omega $ is open to other kinds of functions that segregates features and enhances the performance of the model. $\textbf{Z}_{_{AB}}$ helps in generating the weight-age factor, $\textbf{Z}_{_{A}}$ helps in extracting information from the features  and $\textbf{Z}_{_{B}}$ helps in generating relevant counterparts from another learned feature space like NLP.  The video captioning problem deals with both variational (ResNet) and static (optical flow) features. While creation of a joint sub-space is beneficial, we propose a strategy that regulates separate frame importance selection and fusion at a later stage. This separate frame selection is beneficial for a better performance of the SACT model.

\begin{figure*}[ht]
\vskip 0.2in
\begin{center}
\centerline{\includegraphics[width=2\columnwidth]{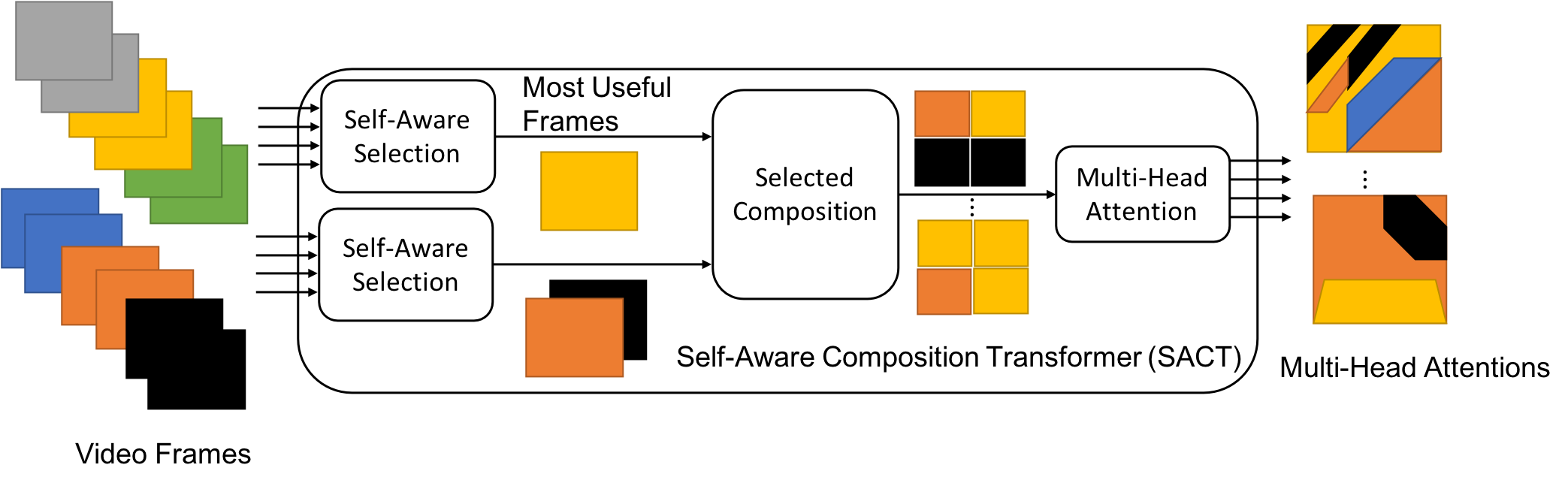}}
\caption{Architectural Details of Self-Segregating Transformer Model for Visual Question Answering.}
\label{fig:an2}
\end{center}
\vskip -0.2in
\end{figure*}

\begin{figure}[ht]
\vskip 0.2in
\begin{center}
\centerline{\includegraphics[width=\columnwidth]{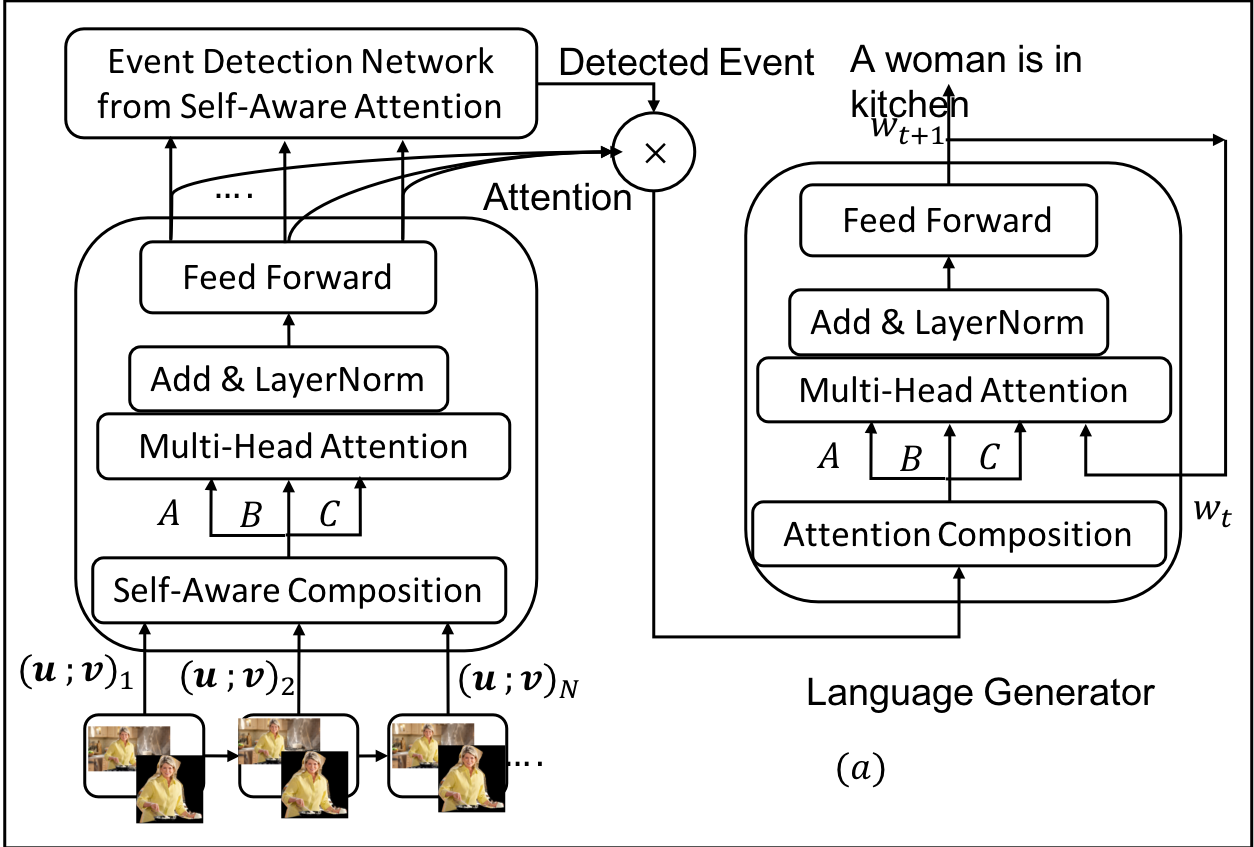}}
\caption{Overall Architecture of the Caption Generated Model with Self-Aware (Joint)  Composition Transformer (SA$_j$CT). }
\label{fig:sact1}
\end{center}
\vskip -0.2in
\end{figure}

\subsection{Self-Aware (Joint) Composition Transformer} 
Self-Aware (Joint) Composition Transformer (SA$_j$CT) includes Self-Aware Composition Attention into the system to create a joint understanding of the two feature spaces. It merges the two spaces, traditionally useful for NLP applications. But, for applications in video attention, submerging the two different feature spaces will approximate and it will not be a good strategy for representation learning for captions. Some of the salient features of the SA$_j$CT include convergence of the two features. But this has limited exploration of the original image  features, the static features (optical flow) having more influence on the attentions. Previous works didn't consider this limitations and hence, in spite of increased dimension, there was no considerable improvement in performance. This strategy can be beneficial for some applications related to the creation of the correlation of the two feature spaces and creating transformations, contributed from them. Applications, like captioning, require feature preference, visual-reasoning and composition. We introduced another strategy, SA$_s$CT, to solve the joint-selection dilemma. 
Mathematically, we can define SA$_j$CT with the following set of equations. 
\begin{equation}
\begin{split}
 f & = \psi_N(\textbf{W}_p\Phi(\textbf{u},\textbf{v}), \textbf{W}_q\Phi(\textbf{u},\textbf{v}), \textbf{W}_r\Phi(\textbf{u},\textbf{v})) \\
   & = [\text{head}_1, \text{head}_2, \ldots, \text{head}_h] \textbf{W}
 \end{split}
\end{equation}
where we define $\text{head}_h$ as, 
\begin{equation}
 \text{head}_h = \psi(\textbf{W}_{p_h}\Phi(\textbf{u},\textbf{v}), \textbf{W}_{q_h}\Phi(\textbf{u},\textbf{v}), \textbf{W}_{r_h}\Phi(\textbf{u},\textbf{v}))
\end{equation}
where we have included $ \psi(.) $ from Equation \ref{eq:SACA1} and Equation \ref{eq:SACA4} from the previous section on Self-Aware Composition Attention. 
Considering feature frames $\{u_1 , \ldots , u_n\}$, Self-Aware (Joint) Composition Transformer can be denoted as, 
\begin{equation} \label{eq:exp1a}
 \begin{split}
  H & = T_1(\{(u_1,v_1) , \ldots , (u_n,v_n)\}) \\
  & = Q_{A_1} (  Q_{S_1} (\{(u_1,v_1) , \ldots , (u_n,v_n)\})  )  
 \end{split}
\end{equation}
\begin{equation}
 R = Q_E (H) 
\end{equation}
\begin{equation}\label{eq:exp1b}
\begin{split}
 L(\{(u_1,v_1) , & \ldots , (u_n,v_n)\}) = L( T_2(H * R) ) \\
 & = L( Q_{A_2} (  Q_{S_2} ( H * R )  ) )
\end{split}
\end{equation}
$ Q_{A_*} (.) $ is attention function, $ Q_S (.) $ is selection of the frames based on self-awareness, $ Q_{A_*} (.) + Q_S (.) $  is transformer function denoted as $T(.)$.  $Q_E (.)$ is the event proposal network and $L(.)$ is language decoder for $\{w_1 , \ldots , w_n\} $.

\subsection{Self-Aware (Separated) Composition Transformer}
In SA$_j$CT, the awareness is propagated on the joint concatenation of the two features existing on different sub-spaces like ResNet and optical flow. Both of these features have individual characteristics. While ResNet features tries to generate the different variational aspects (mainly the changes), optical flow deals with the constant aspects like regional of interest. A joint combination can be difficult to converge or one have high influence than the other. Hence, we introduced Self-Aware (Separated) Composition Transformer (SA$_s$CT).  SA$_s$CT tries to make the model learn to detect useful features independently and then converge on large scale multi-head attention generation. This approach is very important for applications that have different features for generating different aspects of the same detected event. SA$_s$CT feature attentions are much more accomplished (separate identification of usefulness) and will help in enhancing the performance. Figure \ref{fig:sact2} provides a diagram of the SA$_s$CT architecture. Mathematically, we can define (SA$_s$CT) with the following set of equations. 
\begin{equation}
\begin{split}
 f_1 & = \psi_{N_1}(\textbf{W}_{1_p}\Phi(\textbf{u}), \textbf{W}_{1_q}\Phi(\textbf{u}), \textbf{W}_{1_r}\Phi(\textbf{u})) \\
   & = [\text{head}_{1_1}, \text{head}_{1_2}, \ldots, \text{head}_{1_h}] \textbf{W}
 \end{split}
\end{equation}
\begin{equation}
\begin{split}
 f_2 & = \psi_{N_2}(\textbf{W}_{2_p}\Phi(\textbf{v}), \textbf{W}_{2_q}\Phi(\textbf{v}), \textbf{W}_{2_r}\Phi(\textbf{v})) \\
   & = [\text{head}_{2_1}, \text{head}_{2_2}, \ldots, \text{head}_{2_h}] \textbf{W}
 \end{split}
\end{equation}
We have the multi-head attention individuals $\text{head}_h$ for each $h \in \{1,\ldots,h\}$ as,
\begin{equation}
 \text{head}_{{1/2}_h} = \psi(\textbf{W}_{{{1/2}_p}_h}\Phi(\textbf{u}/\textbf{v}), \textbf{W}_{{{1/2}_q}_h}\Phi(\textbf{u}/\textbf{v}), \textbf{W}_{{{1/2}_r}_h}\Phi(\textbf{u}/\textbf{v}))
\end{equation}
Final attention can be defined as the joint selection of the features as the followings,
\begin{equation}
  f = [ f_1 ; f_2 ] = [  ( \text{head}_{1_1};\text{head}_{2_1} ) , \ldots , ( \text{head}_{1_h};\text{head}_{2_h} )  ]
\end{equation}
It is important to note that the individual attention can create sub-space awareness and utilization index can be improved for high dimensional problems. In video captioning application, the typical size of the sequence is $480$ frames, through the event proposal network helped in making some of the contents. 
Considering frames $\{u_1 , \ldots , u_n\}$ and $\{v_1 , \ldots , v_n\}$, Self-Aware (Separated) Composition Transformer Network for our experiment can be denoted as,
\begin{equation}
  H_1 = T_1(\{u_1 , \ldots , u_n\}) = Q_{A_1} (  Q_{S_1} (\{u_1 , \ldots , u_n\})  )  
\end{equation}
\begin{equation}
  H_2 = T_2(\{v_1 , \ldots , v_n\}) = Q_{A_2} (  Q_{S_2} (\{v_1 , \ldots , v_n\})  )  
\end{equation}
\begin{equation}
 R = Q_E ((H_1;H_2)) 
\end{equation}
\begin{equation}
\begin{split}
 L(\{(u_1,v_1) ,  & ..  , (u_n,v_n)\})   = L( T_3((H_1;H_2) * R) )  \\
 & = L( Q_{A_3} (  Q_{S_3} ( (H_1;H_2) * R )  ) )
\end{split}
\end{equation}
The notations are the same as in Equation \ref{eq:exp1a}-\ref{eq:exp1b} and defined in the previous subsection.

\subsection{Event Detector Model}
We used the approach proposed in \cite{zhou2018towards} for the event detection network. The main task of the event detector module is to detect the events that can be narrated through captions. It also helps in identification of the regions that the network is trained to narrate. The events are detected through anchors based on the adjacency of different frames and their relative positions. Different anchor-offset are defined including event proposal score $O_i \in [0, 1]$ for a sequence of $1$ to $N$ and $i \in \{1,\ldots,N\}$. It is supported by other two anchors called center  anchor and length anchor. All these are done using a 1-D temporal convolution network for determination of the proposal. The convolution layer will be sensitive to the positional identification and its relative position for different activities. The center and length offset will capture the events as a high estimated likelihood  for convolution output for a certain moving window. The anchors-offsets adjust the proposed boundaries for the associated locations of the events in a segment. Since the multi-head self-attention layer is used in the prediction of the anchors, it is evident that the adjacency of the frames have high influence on the representation of the features as well. This prevents proper selection of the useful features and also prevents establishment of non-adjacency frames based attention. Self-Awareness strategy will overcome this and will encourage cross-adjacency attention for better representation.

\subsection{Language Generator Transformer}
A masked transformer model is used as a language decoder with masked as \cite{devlin2018bert, zhou2018end} for better training.  It is based on the fact that a masked model can help in significant improvements of language generation model. It also helps in learning better embedding for the word contexts and the positional tensors for language attribute interpretations. Several works with transformer networks \cite{devlin2018bert} have provided instances that the language representation improved with masking, which is one kind of stochastic regularization. But, there is no evidence, whether masking will help in better captioning. Unlike pre-trained model frameworks, our model has to be trained from scratch. Masking helps in preventing the over dependency for the appearance of high posterior likelihood of a word on its previous word. We prefer skipping feedback of the originally generated word with a masked component embedding, provided the model is trained to generate enough content and context from the multi-head attentions.


\section{Experiments, Results \& Discussion} \label{section:results}

\begin{table*}[!ht]
\caption{A Comparative Study of the Self-Aware (Joint) Composition Transformer and Self-Aware (Separated) Composition Transformer Architecture with Existing Works for ActivityNet Data.}
\label{table:an1}
\vskip 0.15in
\begin{center}
\begin{small}
\begin{sc}
\begin{tabular}{|l|c| c| c| c| } 
\hline
Model \& Data & BLEU\_3 & BLEU\_4 & METEOR  \\ 
(ActivityNet Data) & & &   \\\hline\hline
Bi-LSTM+TempoAttn \cite{zhou2018end} & 2.43 &  1.01 & 7.49 \\
End-to-end Masked Transformer \cite{zhou2018end} & 4.76 & 2.23 & 9.56 \\
Masked Transformer \cite{zhou2018end} & 4.47 & 2.14 & 9.43 \\ 
  SA$_s$CT (ours) &  \textbf{5.10} &  \textbf{2.95}  & \textbf{10.30}  \\ 
  SA$_j$CT (ours) & 4.92  & 2.88  &  10.16 \\ \hline 
\end{tabular}
\end{sc}
\end{small}
\end{center}
\vskip -0.1in
\end{table*}

\begin{table*}[!ht]
\caption{A Comparative Study of the Self-Aware (Joint) Composition Transformer and Self-Aware (Separated) Composition Transformer Architecture with Existing Works for  YouCookII Data.}
\label{table:an2}
\vskip 0.15in
\begin{center}
\begin{small}
\begin{sc}
\begin{tabular}{|l|c| c| } 
\hline
Model \& Data & BLEU\_4 & METEOR  \\ 
(YouCookII Data) & &   \\
\hline\hline
  Bi-LSTM+TempoAttn \cite{zhou2018end} & 0.08 & 4.62 \\
  End-to-end Masked Transformer \cite{zhou2018end} & 0.30 & 6.58  \\
  SA$_s$CT (ours) & \textbf{0.48} & \textbf{7.34}  \\ 
  SA$_j$CT (ours) & 0.42 & 6.95  \\ \hline
\end{tabular}
\end{sc}
\end{small}
\end{center}
\vskip -0.1in
\end{table*}

\subsection{Dataset Description}
We have done extensive experiments on different datasets for our proposed model.  Video captioning is highly dependent on event detection, but never guarantee effectiveness due to the large amount of temporal series features and the model has no sense of their individual utility. Instead of depending on capturing the attention through weights, it is more effective to differentiate the frames based on the content and segregate as composition like Multinomial Attention. 
We have utilized this concept for video captioning application on ActivityNet and YouCookII dataset and used the features shared by \cite{zhou2018end}.  It consisted of $2048$ dimension for both ResNet features and optical flow features. The optical features consisted of the main region-of-interest in the video dataset. ActivityNet dataset was part of CVPR $2017$ ActivityNet Video Dense-captioning Challenge. ActivityNet and YouCookII are the two largest datasets with temporal event segments that are annotated and described by natural language sentences. ActivityNet contains $20k$ videos, and each video has $3.65$ events annotated on an average. On the contrary, the YouCookII dataset has $2k$ videos and $7.70$ segments annotated on average. The training/validation/testing data division amounts for ActivityNet and YouCookII are $50\%:25\%:25\%$ and $66\%:23\%:10\%$ respectively. The experimental results for both datasets are being reported on the validation sets in Table \ref{table:an1}. 


\subsection{Model Analysis}
Most of our experiments are based on the masked transformer training strategy introduced by \cite{zhou2018end} and the event detection strategy of \cite{zhou2018towards}.  The novel SACT strategy, when inserted at appropriate positions, helped in improving the performance. However, it is very difficult to get through the benchmark of \cite{zhou2018end}, mainly when the dataset is difficult as ActivityNet and YouCookII. YouCookII is a cooking description dataset and ActivityNet is a sports related commentating dataset. Naturally, it is difficult to predict the center anchor, length anchor and the anchor based proposal offsets for the whole sequence. However, we found that the Self-Aware strategy offered the most beneficial result, when they were introduced at the end of the transformer on a multi-layered network. SACT must be sensitive to detect individual frames (attention for other layers of SACT). With multi-layered SACT, the subsequent layers are sensitive to detect the usefulness of the multi-head attentions. The SACT strategy also helps in learning the representation in  the absence of the uncaptured minute details. 
It is a very difficult to achieve very high BLEU$\_4$ accuracy for video captioning, mainly when the performance is dependent on the event detector to predict the right spots of the frames. Even, the event detector is dependent on the multi-head attention to determine the region of interest frames in the sequence. That means that the representation for event detection and for caption generation must be different. 
In Table \ref{table:an1}, we experimented with ActivityNet and out-performed the previous results. We ran the experiments for 30 epoch for better training of the awareness modules for a end-to-end model with masking framework. Most of the dimensions remaining same as \cite{zhou2018end}, except the learning rate ($=0.0002$) and epochs. Overall attention dimension is 1048, 512 from both image feature and optical features. Masking creates scope of "not-present" situations in the language data to avoid over-fitting and learning a sweet sub-space for the language embedding, consisting of positional and contextual information. Also, it prevents the dependency of the likelihood estimation of word $w_{t}$ less on the $w_{t-1}$ embedding and more on the generated multi-head attention.
In Table \ref{table:an2}, YouCookII was used for our experiment for both the SA$_s$CT and SA$_j$CT model. The cooking dataset is very difficult to train and provide accurate description, but this research work is a good start in the right direction. From the prospect of model performance, SA$_s$CT performed better than SA$_j$CT. This is obvious from the fact that in SA$_s$CT, the selection occurred at a separate level and helped in better attention modeling than SA$_j$CT. However, this conclusion must not be generalized across different domains and applications.

\subsection{Quantitative Comparison Analysis}
Different language metrics like Bleu\_n (n $= 1, 2, 3, 4$), METEOR, ROUGE\_L, and CIDEr-D must be used to evaluate the performance of the generated captions. These statistical methods are standardized in the language generation and language translation research community and are widely used for performance evaluation and comparison purposes. However, it must be mentioned that each of these metrics reflects very limited perspective of the generated captions and, hence, a qualitative evaluation of the generated captions is inevitable. However, BLEU$\_4$ can always be a good judge for determining the seriality of some of the nouns, verbs and language descriptive attributes like adjective and adverbs.

\subsection{Qualitative Analysis} 
Numerical evaluation can never be perfect judge for any AI model, unless its potential is judged through tangible evidences. Also, when it is language generation, normal statistical language metrics can never be the perfect judge for the generated sentences. Hence, it is very important to see the individuals for grammatical (syntactic) significance and conceptual (semantic) clarity. Figures in Supplement material will provide some instances of the generated captions along with details.

\section{Conclusion} \label{section:conclusion}
In this work, we proposed a new concept to engage more refined of feature content and composition of usable attentions for deriving better representation. While previous works concentrated on the utilization of most of the features for attention, reducing the feature space will improve the quality and create better attention composition. With this concept, we defined Self-Aware Composition Transformer architecture with feature understanding at the frame level for video content description. This novel architecture with better content understanding helps with  improved regional proposal and later translates the filtered contents to better captions for videos. Compared to the previous works, this architecture has the advantage of more refined attention to composition and understanding of the video frames. This is a requisite for enhancing the understanding of the machines.  This gradually gives rise to the topologically relevant representations that can be translated to enhanced descriptive captions. This architecture has surpassed many previous works and has produced a new benchmark for ActivityNet Captions and YouCookII datasets.

\ifCLASSOPTIONcaptionsoff
  \newpage
\fi

%








\end{document}